# Adverse Conditions and ASR Techniques for Robust Speech User Interface

Urmila Shrawankar[1], Vilas Thakare[2]

[1] IEEE Student Member, Research Scholar, G H Raisoni College of Engg. Nagpur, INDIA

[2] SGB Amravati University, Amravati, INDIA

## Abstract

The main motivation for Automatic Speech Recognition (ASR) is efficient interfaces to computers, and for the interfaces to be natural and truly useful, it should provide coverage for a large group of users. The purpose of these tasks is to further improve man-machine communication.

ASR systems exhibit unacceptable degradations in performance when the acoustical environments used for training and testing the system are not the same.

The goal of this research is to increase the robustness of the speech recognition systems with respect to changes in the environment. A system can be labeled as environment-independent if the recognition accuracy for a new environment is the same or higher than that obtained when the system is retrained for that environment. Attaining such performance is the dream of the researchers.

This paper elaborates some of the difficulties with Automatic Speech Recognition (ASR). These difficulties are classified into Speakers characteristics and environmental conditions, and tried to suggest some techniques to compensate variations in speech signal.

This paper focuses on the robustness with respect to speakers' variations and changes in the acoustical environment.

We discussed several different external factors that change the environment and physiological differences that affect the performance of a speech recognition system followed by techniques that are helpful to design a robust ASR system.

## 1. Introduction

Speech recognition, is commonly known as automatic speech recognition (ASR), is the process of converting an acoustic signal, captured by a microphone or a telephone, to a text.

The main goal of speech recognition is to get effective ways for mankind to communicate with computers, for example, voice-controlled personal computers. Today's ASR systems are giving considerable compatibility the performance of such systems is far from the perfect system and the research is still gaining on this area.

There are diversified issues concerning the operation of modern ASR systems, such as antiphons which lead to reductions in their efficiency and accuracy to actuate these issues the existence of various forms of variability an articulation in speech [14]. These include Speaker characteristics & variations in acoustic environment,

### 1.1. Speaker Characteristics

A major factors lead to deformities in the performance of ASR is the articulation environment which leads to commotion such deformity in performances can be boisterous caused by physiological and dialectical differences among the speakers leads to diffident isotopic. Evidence of this can be found by comparing the performance of speaker dependent (SD) and speaker independent (SI) ASR systems. A speaker dependent system is trained using data from a single speaker, and euphony in nature. On the other hand, a speaker independent system is trained using data from a large inconsistent group of different speakers for use by speakers that are not necessarily in the training strata. These techniques can be divided into two main categories: speaker adaptation and speaker normalization.

- Speaker adaptation [11] techniques require the existence of a model or Inunit which has already been trained for single or numerous speakers. The goal of such techniques is to tune the parameters of this model to a new lineament.

  Another technique for adapting HMM-based systems is the maximum likelihood linear regression (MLLR) approach which requires a relatively small amount of adaptation lexicon data. This data is used to compute a number of linear transformations which are applied to the colophon means contained in the model.

- Speaker normalization techniques, [12] on the other hand, instead of replacing the model, perform lingual transformations on the speech signal to compensate for speaker variabilities.

Due to the significant variations in the vocal tract length of different speakers, the positions of the formants produced by different speakers can vary. Therefore, a major category of subtle speaker normalization techniques are focused on normalizing the effective vocal tract length across different vocalist. Vocal tract length normalization (VTLN) and augmented state-space acoustic decoder (MATE) perform this by applying a linear warping to the frequency axis of the utterance, normalizing the position of spectral peaks or formants of speech.

## 1.2. Variations in Acoustic Environment

Another major factor that leads to degradations in the performance of ASR systems is the presence of noise in the environment. Such degradations in performance can be due to the mismatch between the conditions in which the systems are trained and the ones in which they are operated.

Some speech enhancement approaches are found really good to deal with unknown noise and filtering such as,
- Spectral Subtraction
- Spectral Normalization

The paper is arranged as section II will explain difficulties with ASR, section III gives Classification of Parameters that affect ASR includes Prosody and Environmental parameters, Section IV describes techniques that are helpful to design a robust speech recognition system and finally section V is the conclusion.

## 2. Difficulties with ASR

The issue of robustness in speech recognition is pregnable of problems. A speech quest may be robust in one environment and yet be impregnable for another and falter less sensitive to noise are not necessarily less sensitive to speaker variability in noise and stress. The main reason for this is that performance of existing recognition systems which may be succinct environment, degrade rapidly in the presence of noise, distortion, and speaker stress.

## 2.1. Reasons for Difficulties in Speech Recognition:

Difficult problem, largely because of many sources of variability [14, 22] associated with signal
- Acoustic realizations of phonemes, highly dependent on context in which they appear.
  - Phonetic variabilities are exemplified by acoustic differences of phoneme
  - At word boundaries, contextual variations can be quite dramatic
- Acoustic variabilities can result from changes in environment as well as in position and characteristics of transducer.

- The sensitivity to the environment (background noise or channel variability), or
- Within-speaker variabilities can result from changes in person's physical and emotional state, speaking rate, gender, vocal effort, regional accents, speaking style, voice quality etc.
- Differences in sociolinguistic background
- Complexity of the human language, the weak representation of grammatical and semantic knowledge.
- Dialect, and vocal tract size and shape can contribute to across-speaker variabilities.
- Background noise

## 2.2. Approaches To Improve Speech Recognizers:

There are a variety of approaches that can be used to improve the robustness of speech recognition. These can be classified into five general areas as follows:
- Pre-processing Techniques : Feature Analysis
  - Voice Activity Detector [7]
- Feature Enhancement : front-end signal processing
  - Estimation and noise detection [8]
  - Echo and reverberation detection
  - Normalization
- Feature Extraction : Extracting coefficients
- Model Adaptation: adapting recognition models to the noisy speaker conditions.
- Training Models: consider alternative training using either noisy data, mismatch between training/test data, or modifications which cause the trained models to be more effective for recognizing noisy speech.

## 3. Classification of Parameters that Affect ASR

Speech Recognition system [16] ebullient in speech processing we regard this as pitch, duration, intensity, voice quality, signal to noise ratio, voice activity detection and strength of Lombard effect these parameters are categorized under two types:
- **Prosody Parameters:** [23] Pitch, duration, intensity and voice quality etc. These parameters are used in speech recognition and especially in the field of speaker characterization. Those systems have to work in general adverse conditions, which leads to the demand of noise robustness for the algorithms estimating the prosodic parameters. Yet it is well known, that most of these parameters are hard to extract from the speech signal, especially under adverse conditions.
- **Environmental Parameters:** the second is the acoustic properties of the environment including the impact on the speaker's voice. As second set we select 'environmental parameters': signal to noise ratio (SNR), voice activity detection (VAD) [6] and strength of Lombard effect (SLE).

Environmental parameters are used in speech recognition and speaker recognition for noise reduction algorithms. Many approaches exist to estimate SNR and VAD from noisy signals.

The SLE parameters, that decreases the performance of speech recognition systems dramatically with Lombard effect.

## 3.1. Prosodic Parameters
**Speaker variability & characteristics** [17] :
the speech signal is non-stationary it  not only convey semantic information (the message) but also a lot of information about the speaker himself like, gender, age, social and regional origin, health and emotional state and its identity.

All speakers have their special voices, due to their unique biological body structure (Vocal Track Autonomy) and personality. The voice is not only different among speakers; there are also wide in variations within single specific voice.

The speaker uniqueness results from a complex combination of physiological and cultural aspects. While finding the variability among speakers through statistical analytic methods found that the first two principal components correspond to the gender and accent respectively. Gender would then appear as the prime factor related to physiological differences, and accent would be one of the most important from the cultural point of view.

The effect of the vocal tract shape on the intrinsic variability of the speech signal between different speakers has play special role in measuring performance of ASR.

**Technical Mythology for Compensation Speaker Variation:**
Techniques for handling speaker variability are mainly divided into:
- speaker independent feature extraction,
- speaker normalization using The different vocal tract length normalization (VTLN) techniques:
    - speaker-dependent formant mapping
    - transformation of the LPC pole modeling
    - frequency warping, either linear or non-linear
- Speaker adaptation : reduce speaker specificities and tends to further reduce the gap between speaker-dependent and speaker-independent ASR by adapting the acoustic models to a particular speaker

## Table 1 : Speaker variations
## (Please See at the end of the paper)

## 3.2. Environmental Parameters
The environmental parameters [19] VAD and SNR play a major role in speech recognition system. Extrinsic variabilities are due to the environment: signal to noise ratio may be high but also variable within short time.

**Environmental variability & characteristics**
- Speech in high noise, with signal-to-noise ratios (SNRs) at or below 0 dB
- Speech in presence of background speech
- Speech in presence of background music
- Speech in highly reverberant environments

**Technical Mythology for Compensation of Environmental Variation [21]**
General techniques such as
- Compensation,
    - Enhancing speech signal [1],
    - Training models on noisy databases,
    - Designing specific models for noise and speech [3],
    - Considering noise as missing information that can be marginalized in a statistical training of models by making hypotheses on the parametric distributions of noise and speech [9].
- Adaptation [5],
- Multiple models,
- Additional acoustic cues and
- More accurate models

## Table 2 : Environmental Variation
## (Please see at the end of paper)

# 4. ASR Techniques

In this section, we review methodologies towards improved ASR analysis/modeling accuracy and resistance towards variability sources.

## 4.1. Front-end Techniques [15]
Feature extraction front-end techniques for the assumption of non-stationary speech signals, high levels of noise, workload task stress, Lombard effect and other variations like,
- Robust features extraction
- The speaker spectral characteristics of speech variability.
- Front-end noise suppression
- Feature compensation techniques for noise reduction.
- Techniques for combining estimation based on different features sets and dimensionality reduction approaches.
- Model adaptation

- Training and Testing in the same conditions.

**Feature Extraction Models [13]:**

- Mel-Frequency Cepstral Coefficient (MFCC) or Perceptual Linear Prediction (PLP) coefficient, are based on some sort of representation of the smoothed spectral envelope, usually estimated over fixed analysis windows. Such analysis is based on the assumption that the speech signal is quasi-stationary over these segment durations.

- A temporal decomposition technique represents the continuous variation of the LPC parameters as a linearly weighted sum of a number of discrete elementary components. These elementary components are designed such that they have the minimum temporal spread (highly localized in time) resulting in superior coding efficiency.

- A segmental HMM [Achan et al identifies waveform samples at the boundaries between glottal pulse periods with applications in pitch estimation and time-scale modifications.

- The amplitude modulation (AM) and the frequency modulation (FM) used to detect the transition point between the two adjoining QSSs. The power of the residual signal normalized by the number of samples in the window (FM). The AM signal modulates a narrow-band carrier signal (specifically, a monochromatic sinusoidal signal).

- Frequency Scales (M-MFCC, ExpoLog),

- Feature Processing (CMN, VCMN, LP-vs-FFT MFCCs),

- Model Adaptation (PMC), and

- Combinations of gender dependent with gender independent models

- Training and Testing (ANN & HMM).

## 4.2. Statistical Models:

It is assume that the "clean" speech signal is first passed through a linear filter with unit sample response, whose output is then corrupted by uncorrelated additive noise to produce the degraded speech signal. Under these circumstances, the goal of compensation is, in effect, to undo the estimated parameters characterizing the unknown additive noise and the unknown linear filter, and to apply the appropriate inverse operation.

The popular approaches of spectral subtraction and homomorphic deconvolution are special cases of this model, in which either additive noise or linear filtering effects are considered in isolation. When the compensation parameters are estimated jointly, the problem becomes a nonlinear one, and can be solved using algorithms such as codeword-dependent cepstral normalization (CDCN) and vector-Taylor series compensation (VTS).

## 4.3. Acoustic Model

The performance of acoustic model is depending on the model matching to the task, which can be obtained through adequate training data and selecting multi-style training.

**Model Compensation**

- MM decomposition, where dynamic time warping was extended to a 3D-array where the additional dimension represents a noise reference and an optimal path has to be found in this 3D-domain. The major problem was the definition of a local Probability for each box.

- Parallel model decomposition (PMC) where clean speech and noise are both modeled by HMM and where the local probabilities are combined at the level of linear spectrum, this implies that only additive noise can be taken into account.

**Adaptation**

- A Maximum Likelihood (ML) criterion.
  Try to maximize the probability of a given sequence of observations; Baum-Welch method gives the result.

- A Maximum a Posteriori [2] (MAP)
  In Bayesian [4,10] statistics, a maximum a posteriori probability (MAP) estimate is a mode of the posterior distribution. The MAP can be used to obtain a point estimate of an unobserved quantity on the basis of empirical data. MAP estimation can therefore be seen as a regularization of ML estimation

## 4.4. Multiple Modeling [10]

Merging too many heterogeneous data in the training corpus makes acoustic models less discriminant. Hence the numerous investigations along multiple modeling, that is the usage of several models for each unit, each model being train from a subset of the training data, defined according to a priori criteria as gender, age, rate-of-speech (ROS) or through automatic clustering procedures. Ideally subsets should contain homogeneous data, and be large enough for making possible a reliable training of the acoustic models. Gender information is one of the most often used criteria.

- Speaking rate affects notably the recognition performances, thus speaking rate dependent models were studied. It was also noticed that speaking rate dependent models are often getting less speaker-independent because the range of speaking rate shown by different speakers is not the same, and that training procedures robust to sparse data need to be used.

- Signal-to-Noise Ratio (SNR) also impacts recognition performances, hence, besides or in addition to noise reduction techniques, SNR-dependent models have been investigated. In multiple sets of models are trained according to several noise masking levels and the model set appropriate for the estimated noise level is selected automatically in recognition phase. On the opposite, in acoustic models composed under various SNR conditions are run in parallel during decoding.

- Multi-speaker models [16]: If models of some of the factors affecting speech variation are known, adaptive training schemes can be developed, avoiding training data sparsity issues that could result from cluster-based techniques. This has been used for instance in the case of VTLN normalization, where a specific estimation of the vocal tract length (VTL) is associated to each speaker of the training data. This allows to build a canonical models based on appropriately normalized data. During recognition, a VTL is estimated in order to be able to normalize the feature stream before recognition. More general normalization schemes have also been investigated, based on associating transforms (mostly linear transforms) to each speaker, or more generally, to different cluster of the training data. This transformation can also be constrained to reside in reducing dimensionality eigen space. A technique for factorization selected transformations back in the canonical model is also proposed in, providing a flexible way of building factor specific models, for instance multi-speaker models within a particular noise environment, or multi-environment models for a particular speaker.

## 4.5. Models for Auxiliary Parameters [20]

Most of speech recognition systems rely on acoustic parameters that represent the speech spectrum, for example cepstral coefficients. However, these features are sensitive to auxiliary information such as pitch, energy, rate-of-speech, etc. the most simple way of using such parameters (pitch and/or voicing) is their direct introduction in the feature vector, along with the cepstral coefficients.

- Pitch has to be taken into account for the recognition of transonic languages. Various coding and normalization schemes of the pitch parameter are generally applied to make it less speaker dependent; the derivative of the pitch is the most useful feature, and pitch tracking and voicing. Pitch, energy and duration have also been used as prosodic parameters in speech recognition systems, or for reducing ambiguity in post-processing steps. Dynamic Bayesian Networks (DBN) offers an integrated formalism for introducing dependence on auxiliary features.

- Speaking rate is another factor that can be taken into account in such a framework. Most experiments deal with limited vocabulary sizes; extension to large vocabulary continuous speech recognition can be achieve through hybrid HMM/BN acoustic modeling.

- TANDEM approach used with pitch, energy or rate of speech. The TANDEM approach transforms the input features into posterior probabilities of sub-word units using artificial neural networks (ANNs), which are then processed to form input features for conventional speech recognition systems.

- Auxiliary Parameters may be used to normalize spectral parameters, for example based on pitch value is used to modify the parameters of the densities (during decoding) through multiple regressions as with pitch and speaking rate.

## 4.6. Compensation for Environmental Degradation in ASR [18]

Speech samples always affected with the additive noise and linear filtering in normal environment, the use of environmental compensation procedures improves the accuracy in Speech recognition system. The compensation procedures include physiologically-motivated signal processing techniques, modification of either the feature vectors of incoming speech or the internal statistics with which speech recognition systems are trained.

Any change in the environment between the training and testing causes degradation in performance. Continued research is required to improve robustness to new speakers, new dialects, and channel or microphone characteristics, Systems that have some ability to adapt to such changes have to be developed

Some speech enhancement algorithms have proved to be especially important in the development of strategies to cope with unknown noise and filtering.

- Spectral subtraction, to compensate for additive noise. In general, spectral subtraction algorithms attempt to estimate the power spectrum of additive noise in the absence of speech, and then subtract that spectral estimate from the power spectrum of the overall input (which normally includes the sum of speech plus noise). Primarily with the goal of avoiding "musical noise" by "over-subtraction" of the noise spectrum. This method is not appropriate for non-stationary noise, The difficulty to detect pauses (non-speech) in low SNR & musical noise effect. Noise cancellation requires the detection of noise to adaptively extract its spectral and statistical parameters. The ability to discriminate speech from noise enables the calibration of noise cancellation algorithms, identify and filter out these noises from the speech signal. Filtering speech with a high order adaptive FIR filter, when no reference to an external noise source is available, Wiener Filtering for stationary input and noise, no noise reference source is required.

- Spectral Normalization, to compensate for the effects of unknown linear filtering. In general, spectral normalization algorithms first attempt to estimate the average power spectra of speech in the training and testing domains, and then apply the linear filter to the testing speech to "best" convert its spectrum to that of the training speech. Improvements and extensions of spectral subtraction and spectral normalization algorithms

**This section explains some complementary approaches to robust recognition based on initial signal processing techniques.**

**Approaches to Environmental Compensation**
These approaches are grouped as per the effects of noise and filtering.
- Empirical compensation by direct cepstral comparison,
- Model-based compensation, and
- Compensation via cepstral high-pass filtering.

**Empirical Compensation** by direct cepstral comparison is totally data driven, and requires a "stereo" database that contains time-aligned samples of speech that had been simultaneously recorded in the training environment and in representative testing environments. The success of empirical compensation approaches depends on the extent to which the putative testing environments used to develop the parameters of the compensation algorithm are in fact representative of the actual testing environment.

**Empirical Compensation: RATZ and STAR**
- The RATZ algorithm modifies the cepstral vectors of incoming speech,
- The STAR algorithm modifies the internal statistical models used by the recognition system.
- RATZ and STAR have a similar conceptual framework.
- RATZ can be considered to be a generalization of algorithms like MFCDCN.
- STAR can be considered to be an extension of the codebook adaptation algorithms.
- RATZ and STAR both assume that the probability density function for clean speech can be characterized as a mixture density. Where the mixture coefficients are fixed for the case of RATZ, and assumed to vary as a function of time to represent the Markov transitional probabilities for the case of STAR.
- Environmental compensation is introduced by modifying the means and variances of the probability density functions.

**Model-based compensation** assumes a structural model of environmental. Compensation is then provided by applying the appropriate inverse operations. The success of model-based approaches depends on the extent to which the model of degradation used in the compensation process accurately describes the true nature of the degradation to which the speech had been subjected.

**Model-Based Compensation: VTS and VPS**
- The Vector Taylor Series (VTS) and Vector Polynomial Expansion (VTS) algorithms that develop series approximations to the nonlinear environment function.
- The VTS algorithm approximates the environment function using the first several terms of its Taylor series, where is the vector function evaluated at a particular vector point.
- Similarly, represents the matrix derivative of the vector function at a particular vector point. The higher order terms of the Taylor series involve higher order derivatives resulting in tensors.
- The Taylor expansion is exact everywhere when the order of the Taylor series is infinite.
- VPS approach replaces the Taylor series expansion used in VTS with a more general approach to approximating the environment function.
- VPS is shown to provide a more accurate approximation to the environment function than VTS.
- VPS provided somewhat better recognition accuracy compared to VTS, and at a reduced computational cost.
- It is expected that the difference in error rates between VPS and VTS will increase when implementations of these algorithms that modify the internal statistical models are completed.

Compensation by high-pass filtering implies removal of the steady-state components of the cepstral vector. The amount of compensation provided by high-pass filtering is more limited than the compensation provided by the two other types of approaches, but the procedures employed are simple and effective that they should be included in virtually every current speech recognition system.

**Cepstral High-Pass Filtering: RASTA and CMN**
- In Relative Spectral Processing or RASTA processing, a high-pass (or band-pass) filter is applied to a log-spectral representation or cepstral representation of speech.
- Cepstral mean normalization (CMN) is an alternate way to high-pass filter cepstral coefficients.
- High-pass filtering in CMN is accomplished by subtracting the short-term average of cepstral vectors from the incoming cepstral coefficients.
- RASTA and CMN are effective in compensating for the effects of unknown linear filtering in the absence of additive noise because under these circumstances the ideal cepstral compensation vector is a constant that is independent of SNR and VQ cluster identity. Such a compensation vector is, in fact, equal to the long-term average difference between all cepstra of speech in the training and testing environments.
- The high-pass nature of both the RASTA and CMN filters forces the average values of cepstral coefficients to be zero in the training and testing environments individually, which, i, implies that the average cepstra in the two environments are equal to each other.

- Cepstral high-pass filtering can also be thought of as a degenerate case of compensation based on direct cepstral comparison.

**Joint compensation for the effects of noise and filtering has proceeded in two phases.**

- In the initial phase concerned with understanding the basic properties of the environment function and with the development of compensation procedures that were relatively simple but that provided significant improvements in recognition accuracy compared to the accuracy that could be obtained from independent compensation for the effects of noise and filtering.

- During the second phase of algorithm development focused on the development of algorithms that could achieve greater recognition accuracy under the most arduous conditions through the use of more accurate mathematical characterizations of the effects of noise and filtering.

## Table 1 : Speaker Variations

| Reason for Variation | Effect of Variation | A General Technique for handling Speech variation |
|---|---|---|
| Anatomy of vocal tract | The power spectral density of speech varies over time according to the glottal signal and the configuration of the speech articulators. | ▪ Compensation and invariance (Normalization)<br>▪ Vocal Tract Length Normalization (VTLN)<br>▪ Hidden Markov Models (HMMs), as a sequence of stationary random regimes |
| Realization | Speaker can not produce the same acoustic wave for the same word if he/she pronounced over and over again | ▪ Auto corrélation technique<br>▪ Clustering techniques<br>▪ Speeker Adaption model |
| Ambiguity : Homophones Ambiguity : Word boundary | ● Words that sound the same, but have different orthography<br>● Multiple ways of grouping phones into words. | ▪ Language Model<br>▪ Use of multiple acoustic models associated to large groups of pronunciation variants (lexical level) speakers |
| The sex of the speaker | In general Women have shorter vocal tract than men.<br>The fundamental tone of women's voices is roughly two times higher than men's | ▪ Vocal Tract Length Normalization (VTLN)<br>▪ Cepstral Mean subtraction (CMS)<br>▪ Mean & Variance Normalization (MVN)<br>▪ Spectral Normalization |
| Speaking rate & Speaking style | The spectral effects of speech rate variations. | ▪ Speech rate estimator<br>▪ The evaluation of the frequency of phonemes or syllables in a sentence<br>▪ Normalization by dividing the measured phone duration by the average duration of the underlying phone |
| Regional and Social Dialects | Dialects are group related variation within a language.<br>Regional dialect involves features of pronunciation, vocabulary and grammar which differ according to the geographical area.<br>Social dialects are distinguished by features of pronunciation, vocabulary and grammar according to the social group of the speaker. | ▪ Consider dialects as another language in ASR, due to the large differences between two dialects. |
| Amount of data and search space | The quality of the speech signal decreases with a lower sampling rate, resulting in incorrect analysis.<br>Minimizing lexicon (set of words) causes out-of-vocabulary | ▪ Use of large vocabulary |
| Foreign and | Variations in speaker accent degrade the | ▪ Select an appropriate language model or adapt to |

| Regional Accents | performance of speech recognition systems that fails to recognize target language. | the accent/speaker<br>▪ Recognizer should train on target language.<br>▪ Use of multiple acoustic models associated to large groups of pronunciation variants (lexical level) speakers.<br>▪ Adapt of Multilingual phone models |
|---|---|---|
| Age | • The difference in vocal tract size results in a non-linear increase of the formant frequencies.<br>• Larger spectral and supra-segmental variations and wider variability in formant locations and fundamental frequencies in the speech signal. | ▪ Use larges size of the pronunciation dictionary, corpora for children and adults<br>▪ Selection of language models which are customized for children speech & physiological besides.<br>▪ Adapting the acoustic features of children speech to match that of acoustic models trained from adult speech.<br>▪ Use of vocal tract length normalization (VTLN) and spectral normalization approaches |
| Emotions | Emotions in speech recognition is concentrated on attempting to classify a "stressed" speech signal into its correct emotion category.<br>Intrinsic variabilities: loud, soft, Lombard, fast, angry, scared; and noise. | ▪ Improved front-end processing, feature extraction methods for the recognition of stressed and non-stressed speech simultaneously.<br>▪ Improved back-end processing or robust recognition measures.<br>▪ Improved training methods: Multi-style training and simulated stress token generation. |
| Dis-fluencies in speech | False starts, Repetitions, Hesitations and filled pauses, Slips of the tongue etc. | ▪ Spectral Subtraction methods<br>▪ Improved feature extraction methods<br>▪ Appropriate training model |

## Table 2 : Environmental Variation

| Reason for Variation | Effect of Variation | Technique for handling Speech variation |
|---|---|---|
| Noise | Unwanted information in the speech signal like voices in the background that corrupts the quality of speech signal and degrades the performance of ASR system. | ▪ Spectral Subtraction Method<br>▪ Noise Estimation, Cancellation approaches and filters<br>▪ A high order adaptive FIR filter. When no reference to an external noise source is available,<br>▪ Wiener Filtering for stationary input and noise, no noise reference source is required.<br>▪ SPLICE algorithm works on spectral representation<br>▪ ALGONQUIN algorithm works on log-spectra |
| Echo effect | The speech signal bounced on some surrounding object, and that arrives in the microphone a few milliseconds later.<br>This echo effect adds with original speech signal, and difficult to get clean original speech. | ▪ Echo Cancellation Algorithms<br>▪ Least-Mean-Square (LMS) and<br>▪ Normalized LMS (NLMS)<br>▪ Approach for Echo Cancellation<br>▪ Minimum Statistics (MS)<br>▪ Eigenvalue Decomposition<br>▪ Fourier Transform (DFT)<br>▪ State-Space Model<br>▪ Vector Taylor Series (VTS)<br>▪ ALGONQUIN Method<br>▪ Time-Variant Estimate |
| | | ▪ Switching Linear Dynamic Model (SLDM)<br>▪ Bayesian Estimation Framework<br>▪ Random-Walk State Model |
| Reverberation | If the place in which the speech signal has been produced is strongly echoing, then this may give raise to a phenomenon called reverberation, which may last even as long as seconds.<br>The original speech signals mask with echoing. | ▪ Additive noise filtering algorithms<br>▪ Adaptive Schemes<br>▪ Proportionate Schemes<br>▪ Proportionate Adaptive Filters<br>▪ Block-Based Combination<br>▪ Combination Schemes<br>▪ Subband Adaptive Filtering<br>▪ Uniform Over-Sampled DFT Filter Banks<br>▪ Subband Over-Sampled DFT filter banks (FB)<br>▪ Time-Domain Considerations<br>▪ Volterra filters<br>▪ Proportionate-Type Algorithms<br>▪ Sparseness-Controlled Algorithms |
| Channel Variability | The noise that changes over time, and different kinds of microphones and everything else that affects the content of the acoustic wave from the speaker to the discrete representation in a computer. | ▪ Cepstral Mean Subtraction<br>▪ The RASTA filtering of spectral trajectories. |
| Convolution Noise | Speech signal quality degradations due to the channel come from its slowly varying spectral properties (or impulse response). | ▪ Averaging speech features (Cepstral Mean Subtraction)<br>▪ Evaluating the impulse response as missing data and combined with additive noise reduction.<br>▪ Low pass filtering, by removing Cepstral mean from all feature vectors of the utterance. |
| Lombard effect | Due to noisy environments acoustic correlates in the speech signal. But to quantify this effect no specification is known. | ▪ Additive noise filtering algorithms<br>▪ Applying Low pass and High pass filters |
| Physical Stress<br>The force environment, Auditory distraction,<br>Thermal environment,<br>Personal equipment.<br>Emotional Stress<br>Task load,<br>Mental fatigue,<br>Mission anxieties and Background anxieties. | The noise can be considered stationary during a vocal command, but from one vocal command to another, its characteristics can change.<br>. | ▪ Applying suitable feature extraction method like LPCC, MFCC<br>▪ Noise estimation and cancellation algorithms.<br>▪ Noise cancellation to be performed by Wiener type Filtering |

## 5. Discussion & Conclusion

ASR is a challenging task. In this paper, we have addressed some of the difficulties of speech recognition, the most problematic issues being the large search space and the strong variability, this covers accent, speaking rate & style, regional and social dialects, speaker physiology, age, emotions etc.

This paper covered the different causes of acoustical and environmental variability. There are some attributes of the environment that remain relatively constant through the course of an utterance such as the recording equipment, the amount of room reverberation, and the acoustical characteristics of the particular speaker using the system. Other factors, like the noise and signal levels, will be assumed to vary slowly compared to the rate at which speech changes.

Conventional techniques that compensate for the effects of additive noise and linear filtering of speech sounds can provide substantial improvement in recognition accuracy when the cause of the acoustical degradation is quasi-stationary. The recognition of speech at lower SNRs, and especially speech in the presence of transient sources of interference including especially background speech and background music remain essentially unsolved problems at present.

Some techniques are explained for Environment compensation, which remove speech variabilities due to environment and channel characteristics, speaker normalization techniques, which remove variabilities due to speaker characteristics, and discriminant feature space-transformation techniques, which are aimed at increasing the class discrimination of the speech data.

Finally, the paper proposed an overview of general techniques for better handling intrinsic and extrinsic variation sources in ASR, mostly tackling the speech analysis and acoustic modeling aspect.